\documentclass[letterpaper, 10 pt, conference]{ieeeconf}  

\IEEEoverridecommandlockouts                              

\usepackage{cite}
\usepackage{amsmath,amssymb,amsfonts}
\usepackage{algorithmic}
\usepackage{graphicx}
\usepackage{textcomp}
\usepackage{xcolor}

\title{\LARGE \bf
Concepts for End-to-end Augmented Reality based Human-Robot Interaction Systems
}

\author{David Puljiz$^{1}$ and Bj\"orn Hein$^{1,2}$
\thanks{$^{1}$Intelligent Process Automation and Robotics Lab, Karlsruhe Institute of Technology, Karlsruhe, Germany {\tt\small david.puljiz@kit.edu, bjoern.hein@kit.ed}}%
\thanks{$^{2}$Karlsruhe University of Applied Sciences, Karlsruhe, Germany}%
}

\begin{document}

\maketitle
\thispagestyle{empty}
\pagestyle{empty}

\begin{abstract}

The field of Augmented Reality (AR) based Human Robot Interaction (HRI) has progressed significantly since its inception more than two decades ago. With more advanced devices, particularly head-mounted displays (HMD), freely available programming environments and better connectivity, the possible application space expanded significantly. Here we present concepts and systems currently being developed at our lab to enable a truly end-to-end application of AR in HRI, from setting up the working environment of the robot, through programming and finally interaction with the programmed robot. Relevant papers by other authors will also be overviewed. We demonstrate the use of such technologies with systems not inherently designed to be collaborative, namely industrial manipulators. By trying to make such industrial systems easily-installable, collaborative and interactive, the vision of universal robot co-workers can be pushed one step closer to reality. The main goal of the paper is to provide a short overview of the capabilities of HMD-based HRI to researchers unfamiliar with the concepts. For researchers already using such techniques, the hope is to perhaps introduce some new ideas and to broaden the field of research.        

\end{abstract}

\section{Introduction}
Making robots ubiquitous in all industries, from large factories to small enterprises, will require considerable effort and use of new technologies and paradigms. One such technology is AR. Although not new in the field of robotics - indeed the first systems combining the two emerged in the early nineties - the newer generations of devices considerably expand the capabilities and use cases of such systems. These devices include Microsoft's HoloLens and the upcoming HoloLens 2, Magic Leap's Magic Leap One, the Meta's Meta 2 etc. \par

AR can be used to facilitate robot set-up, programming and allow for more intuitive collaboration. It can also be of use in imitation learning, both as a vector to gather data, and to influence the process of learning itself. Furthermore the vast array of sensors and localisation capabilities present on HMDs allow for a paradigm shift - instead of relying on an external sensor suite to track the human coworkers, display information and so on, one can outsource these capabilities to the human coworker through wearables, such as HMDs, allowing simpler, cheaper and reconfigurable robot cells. \par

In this paper we address the use of HMDs to provide the mentioned capabilities. The structure of the paper will be organised as follows. In the next section, Section \ref{sec:sys_comp}, we present the tools and hardware used in development, as well as the capabilities an HMD must have to use similar concepts and algorithms. We proceed to present the basic building blocks of an end-to-end system - referencing (\ref{sec:ref}); robot cell set-up (\ref{sec:cell}); programming (\ref{sec:prog}); HRI in a working system (\ref{sec:int}; and imitation robot learning methods (\ref{sec:lear}). Finally a short conclusion is presented in \ref{sec:conc}.   


\section{System components}
\label{sec:sys_comp}

In this section concepts and research on particular parts of the end-to-end system will be presented. Currently our research, as in most other groups, uses the Microsoft HoloLens as the HMD of choice. The vast majority of methods, however, are also applicable to any HMD device that can: \textbf{(1)} self-localise in an environment; \textbf{(2)} produce a map/mesh of the environment; \textbf{(3)} track hands and gestures; \textbf{(4)} have at least a single RGB camera accessible. These capabilities are ubiquitous in modern HMDs and will certainly be available and even expended in next generation devices.  \par

For the robot control and communication, the Robotic Operating System (ROS) is used, as well as MoveIt! for path planning. For communication between ROS and the HoloLens, rosbridge together with ROS\# toolkit by Siemens is used. For HoloLens application development we use Unity3D. Tests and interactions are carried out on industrial manipulators such as KUKA's KR-5 and KR-16.\par

\subsection{Referencing}
\label{sec:ref}

The first step in any AR-based HRI system is to establish a robust transformation between virtual objects in the HMD's coordinate system and the robot's coordinate system. We call this referencing. This allows not just proper overlay of information over relevant objects, but also allows virtual objects to convey information to the robot e.g. setting up virtual waypoints for the robot's trajectory. Presently most system achieve this either through use of visual markers or through manual referencing of coordinates where the user positions the virtual robot over the real one. AR software packages such as \textit{Vuforia} and \textit{VisionLib} are extending the approach to include 3D object matching, either from 2D images or from the 3D spatial mesh generated by HMDs. We are taking similar approaches, extending the current referencing methods to include 3D object matching. We proposed a semi-automatic method, where user input is required, an automatic one where no input is required, and finally a continuous automatic one which updates the referencing during the interaction. \par

In the semi-automatic method, the user gives a rough initial guess through a seed hologram which is then processed by a standard registration algorithm, ICP \cite{icp} or Super4PCS \cite{super4pcs}.  The registration algorithm refines the guess and provides a more precise referencing. In Fig.~\ref{fig:hand_track} the spatial mesh used for referencing, the seed hologram, and the result of the referencing algorithm can be seen. \par

The automatic method doesn't require user input, thus the first guess should be performed autonomously. We dilate the bounding box of the robot and calculate descriptors - OUR-CVFH \cite{aldoma2012our}, SHOT \cite{tombari2010unique} or a custom-made descriptor consisting of the centroid,  point cloud eigenvalues, eigenvectors and the number of points in n vertical slices. We then slide the bounding box through the scene and match descriptors. The scene box with the best match is selected and one of the previous registration algorithms applied to the selected bounding box to get the final transform. Although each descriptor performs well with certain robots and surrounding environments in our tests, none of them are general, thus more research is required. \par

The previous two methods perform referencing only once, which leaves the system vulnerable if the HMD loses localisation. This usually implies that the HMD will move its world coordinate system to its current location, invalidating the original referencing. Although this can be detected and the user prompted to re-reference, it can be inefficient and annoying, especially in the interaction stage. To combat this, a continuous automatic referencing method is also being developed. The method uses the stream from two of the HoloLens' environmental cameras to form a stereo pair. The frames are captured and compared to previous frames to extract the optical flow. The part of the optical flow that coincides with the ego-motion of the HoloLens is subtracted, leaving only the optical flow of moving objects- in this case the robot. Using the baseline and parameters of the two cameras, a point cloud of the moving parts can be created and compared to the robot model transformed with the current joint states. This method is still heavily in development and results are pending. \par

There are many other methods that can be used for referencing, indeed the problem is equivalent to an autonomous robot trying to find objects in the scene. Other machine vision algorithms based on deep learning such as the bounding box generating Faster-RCNN \cite{Ren2015FasterRCNN}, FCNs \cite{Long2015FCNN} for semantic segmentation or end-to-end approaches such as PoseNet \cite{kendall2015posenet} can be used. Alternatively classical approaches using 3D descriptors, filters and registration algorithms can also be used. For an overview of referencing in AR-based HRI and a more in-depth discussion of the algorithms presented please see \cite{puljiz2019referencing}.\par


\subsection{Cell Set-up}
\label{sec:cell}

The second step after obtaining a precise coordinate transform is to set-up the robot's working environment, meaning setting up safety zones and mapping obstacles. Robot cells are usually planed offline with CAD data, which in turn defines safety zones. The cell is then built, the safety zones defined and uploaded to the robot controller, and then thoroughly tested. Such an approach is time consuming and not applicable in truly flexible, semi-structured production environments. \par

In our approach the in-built spatial mapping capabilities of the HMD is leveraged to create a map of the manipulator's surroundings. The mesh is converted to a point cloud, filtered, points belonging to the manipulator removed and finally converted to an octomap that covers all the obstacles in the environment. The user can edit this octomap and set up safety zones immediately, visualising the safety zones and octomap in the real working environment. The octomap and the safety zones are then used to define the planning scene in the MoveIt! planning framework. \par

\begin{figure}[t]
    \includegraphics[width=0.48\textwidth]{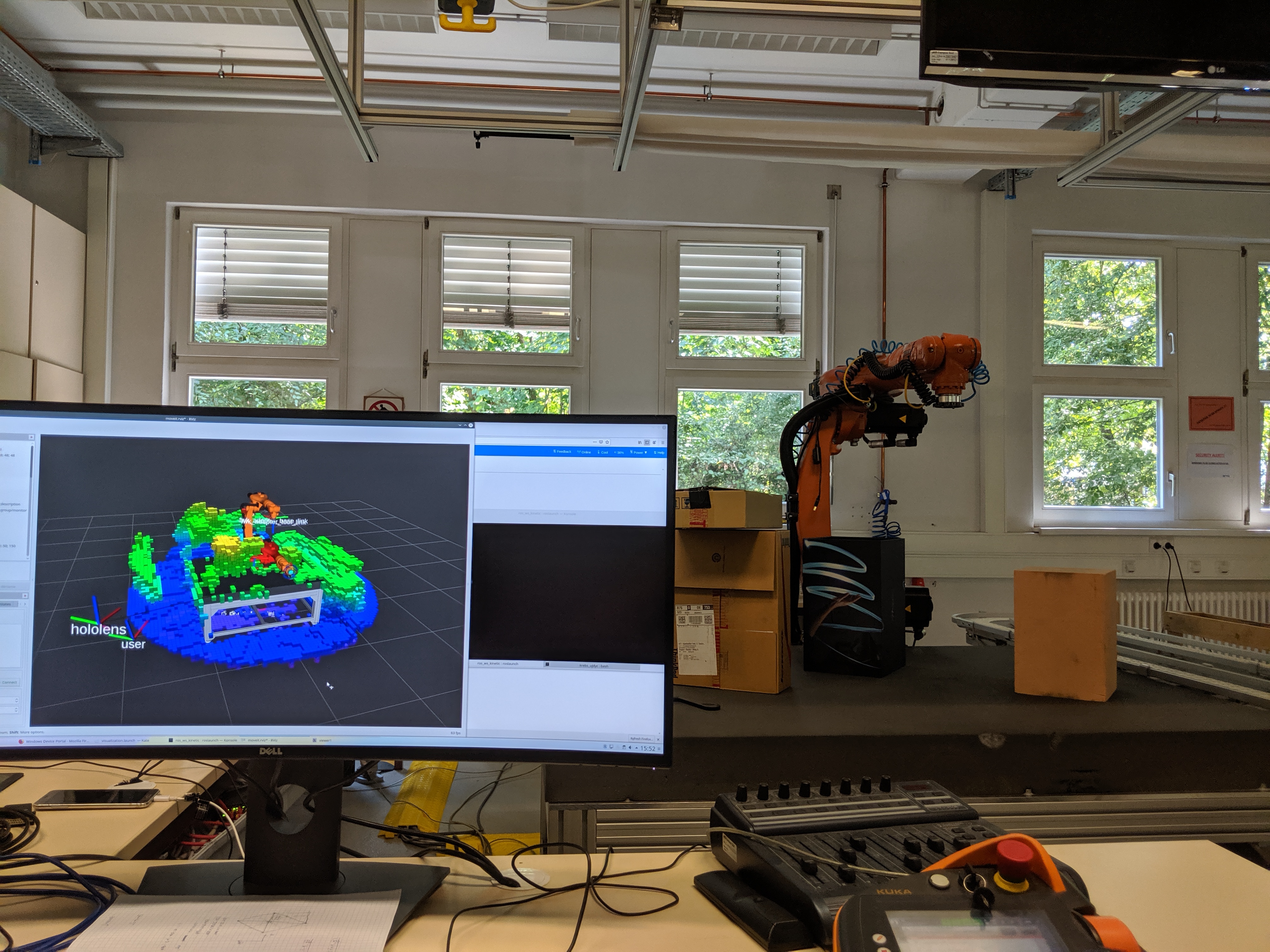}
    \caption{The cell setup process. On the screen the octomap generated from the Hololens' spatial mesh is visible. One can also see the the Hololens' global coordinate system in the robot's coordinate system as well as the current position of the device. In the background the real robot with the obstacles is visible.}
    \label{fig:cell}
\end{figure}

In Fig.~\ref{fig:cell} one can see the octomap in Rviz. The octomap is generated from the processed spatial mesh of the HoloLens, such as the one visible in Fig.~\ref{fig:hand_track}. The octomap is then used as the planning scene for the MoveIt! path planning framework. In Fig.~\ref{fig:cell} one can also see on the monitor that the robot in the goal state is in red as it collides with the obstacles in the environment. \par  

\subsection{Programming}
\label{sec:prog}

\begin{figure*}[ht]
\centering
    \includegraphics[width=0.53\textwidth]{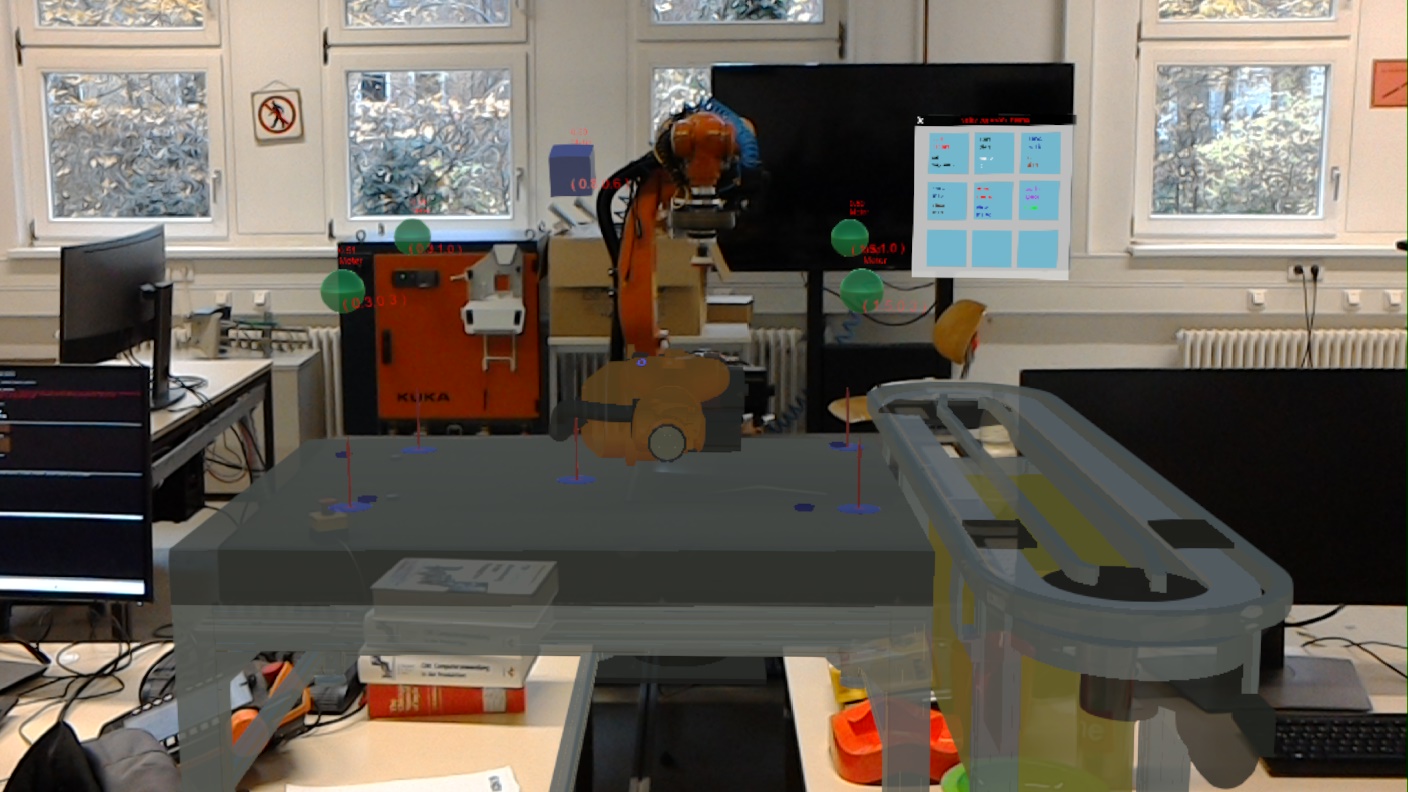}
    \includegraphics[width=0.46\textwidth]{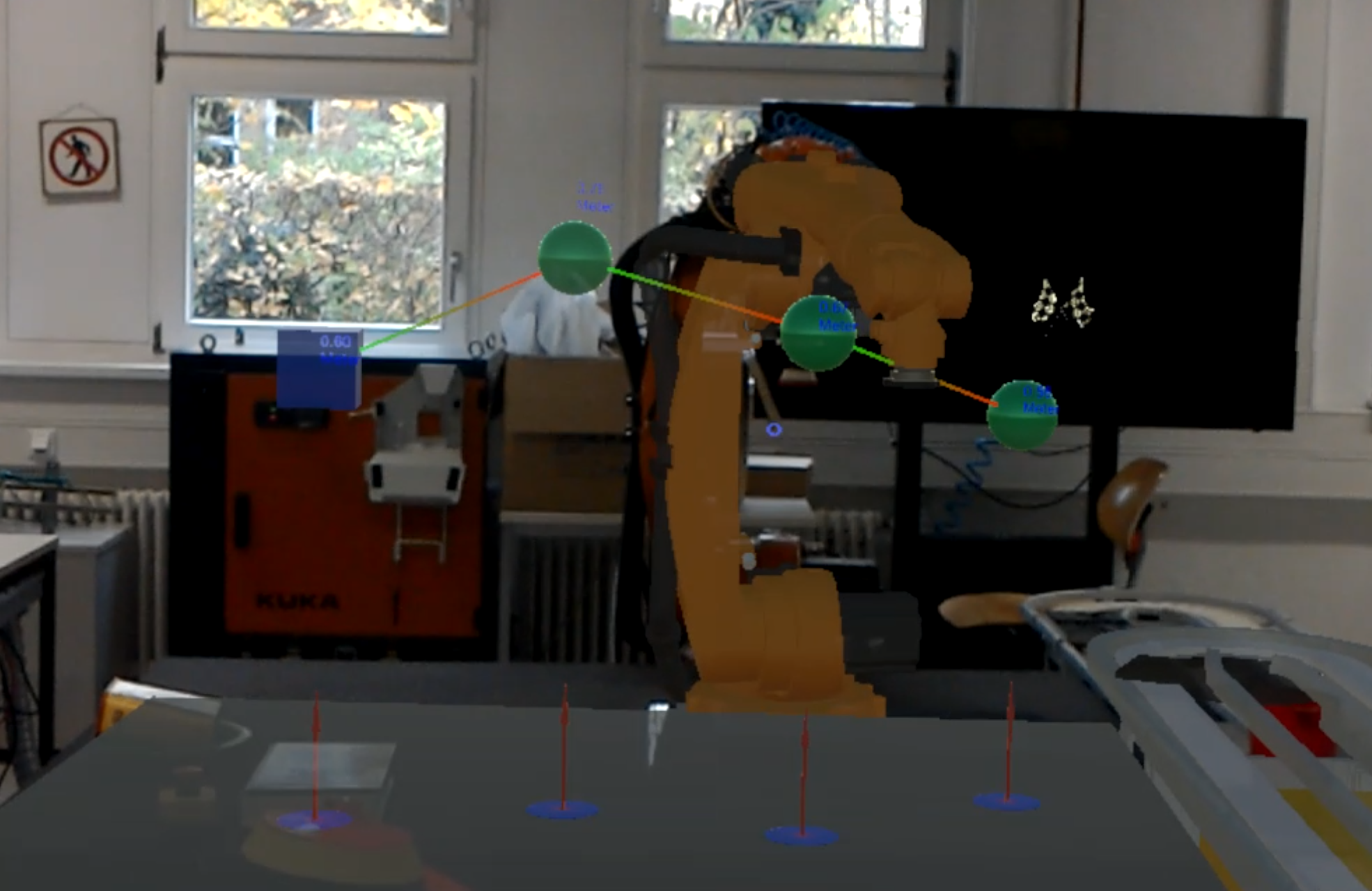}
    \caption[]{The robot programming setup. Left: The waypoints, their position in the coordinate system of the robot, their projection to the work surface as well as the menu are visible. Right: The path and the virtual execution by the robot (different set of waypoints) }
    \label{fig:prog}
\end{figure*}

AR also extends the possibilities in robot programming by allowing waypoint placement and virtual trajectory execution inside the real robot cell. Similar to \cite{Quintero2018ARProg} we implement a waypoint based robot programming system, including free-space wayoints, as well as surface waypoints which are embedded into the spatial mesh. We extend the approach to include visualisation of the exact location of waypoints in the robot coordinate system, the projection of waypoints onto the spatial mesh of the robot cell (useful e.g. when trying to position the robot above a certain point in the workspace) and the integration of the framework with the MoveIt path planning software (instead of calculating B-splines as in \cite{Quintero2018ARProg}). The framework allows to visualise the trajectory, show the virtual robot performing the trajectory before actual execution and, if the trajectory is not satisfying, adding, deleting and moving waypoints to create a new trajectory. Combined with the previous step this allows us to plan custom, collision free trajectories in a cell with static obstacles. \par

As an additional modality, we added hand guidance, such as found on collaborative robots, through the use of the hand tracking capabilities of the HMD. This approach doesn't require any Joint-Torque or other sensors on the robot, doesn't require any direct force to move the robot and isn't controller or robot dependent, indeed working with manipulators with any number of joints. The movements of the hands near a specific links get transformed to the joint values of the parent joint, while the rest of the movement that the joint cannot perform is propagated down the kinematic chain. The framework also allows the user to both guide the end-effector (by invoking inverse kinematic calculation from the path planning software) or move specific joints themselves (by the method described above), thus allowing to make use of any redundant joints. On the downside moving specific joints requires more effort and time on the side of the user. Thus dragging the end-effector should be used for most of the path planning while the joint movement propagation should be used to refine robot poses at specific points for robots with redundant joints. More details on the approach can be found in \cite{puljiz2019handguidance}. \par

There are also other possibilities of inputting desired trajectories. A commonly proposed approach involves the use of input devices, such as pens \cite{gaschler2014intuitive}. Although classically tracked by external tracking sensors, this could now be done using the HMD itself. By using the hand tracking capabilities of HMDs, one can also trace trajectories by hand motion. Integrating voice commands for specific higher level tasks can extend the programming options. Finally one can also specify forces when in contact with the environment regardless weather the waypoint method or some other is used, such as presented in \cite{Quintero2018ARProg}. The HMD-based AR methods are  particularly attractive and useful since they offer numerous input modalities, virtual trajectory execution and programming while inside the real robot cell. 


\subsection{Interaction}
\label{sec:int}

\begin{figure}[h]
    \includegraphics[width=0.48\textwidth]{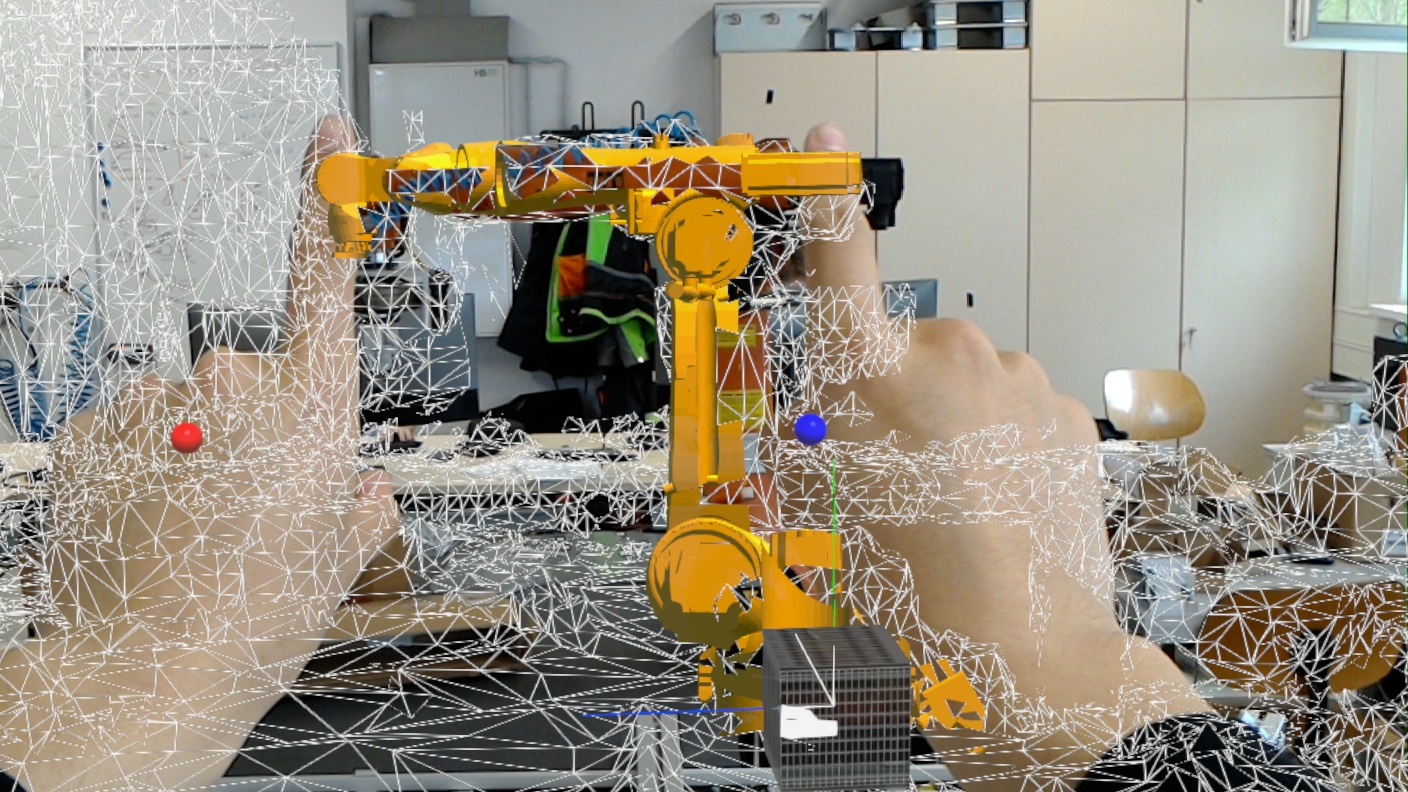}
    \caption{Referencing and hand detection of HoloLens. Here the spatial mesh - used in several stages of the system - is visible. The seed hologram (cube) used as the initial guess for the semi-automatic referencing algorithm is also visible. The virtual robot is overlayed over the real one. Additionally the output of the hand tracking is displayed with the red and blue spheres for the left and right hand respectfully.}
    \label{fig:hand_track}
\end{figure}

Once the manipulator is programmed, methods to interact with human coworkers while the robot performs the task need to be developed. The most basic requirement is knowing the pose of the human to prevent dangerous collisions. We are developing a method based on an HMD and wrist mounted inertial measurement units (IMU). The method doesn't require external tracking sensors or long calibration. It reduces the IMU drift through sensor fusion of the localisation and hand tracking data of the HMD on one side, and the measurements of the IMUs on the other. While the hands are outside the field of view of the HMD's sensors, the filtered data of the IMUs are used for odometry. Here we apply either Kalman filtering or a combination of low and high pass filters. Although this still accumulates drift, the sensor fusion with the HMD's hand tracking capabilities works to reset the drift, allowing for continual use. Thus the pose of the head and both hands in the robot coordinate system can be known. The HoloLens hand tracking capability can be seen in Fig.~\ref{fig:hand_track}. \par

Another modality to facilitate intuitive HRI is displaying robot goals and allowing robots to reference certain objects such as in \cite{weng2019refe}. This allows human coworkers to have much better situational awareness which increases trust and comfort while working with robots. Such a collaborative system was shown in \cite{chakraborti2017}. \par

\subsection{Learning}
\label{sec:lear}

The sensor suit of a modern HMD can provide a significant amount of data to imitation learning algorithms. In particular data on object grasping, manipulation and the sequence of actions needed to perform a task. To achieve this a robust hand skeleton tracking algorithm that is able to cope with contacts between the hand, the objects and the environment is quite important. Such an algorithm was proposed in \cite{GANeratedHands_CVPR2018}. Additionally, estimation of contact points, contact forces, and the deformation of objects during manipulation by the human can provide a plethora of useful data for imitation learning. This is usually achieve by bulky sensor gloves not adapted for use in real working environments. Deep learning algorithms, such as the one proposed in  \cite{Zhang2019InteractionFusion}, can take depth and camera data from the HMD as input, and output object deformation. This would make it possible to gather data in real working environments provided the user is wearing an HMD, which should become increasingly common. \par

Although currently not a strong focus of research in general, applications to use the HoloLens for knowledge patching in temporal and-or graphs \cite{liu2018interactive} and visualising and applying constraints in constrained learning by demonstration \cite{luebbers2019augmented} have already been demonstrated.


\section{Conclusion}
\label{sec:conc}

In this paper we described concepts and algorithms for HMDs to allow end-to-end interaction with general robotic manipulators, from setting up the working environment, through programming and interaction, to it's use in imitation learning. This paper provides a brief overview of the major topics currently being researched in the field of AR-based HRI. The main goal is to show how, by leveraging the new generation of HMDs, many problems that prevent a flexible, collaborative robot work cell can be solved or at least mitigated. \par

Still it is not a catch-all solution. To implement it in real production environments the precision of referencing and path programming needs to be tested, the safety certification aspects assessed, failure cases identified, ergonomy studied etc. In the end other technologies and paradigms will most likely be needed for a truly flexible and safe collaborative system. However, it would seem that HMD-based AR will be a major building block of HRI systems in the future. The research is still relatively in it's infancy and there are still plenty of ways to expand and improve current AR-based HRI paradigms.   

\section*{ACKNOWLEDGMENT}

This work has been supported from the European Union’s Horizon 2020 research and innovation programme under grant agreement No 688117 “Safe human-robot interaction in logistic applications for highly flexible warehouses (SafeLog)”. \par

\bibliographystyle{IEEEtran}
\bibliography{bibliography}

\begin{thebibliography}{10}
\providecommand{\url}[1]{#1}
\csname url@samestyle\endcsname
\providecommand{\newblock}{\relax}
\providecommand{\bibinfo}[2]{#2}
\providecommand{\BIBentrySTDinterwordspacing}{\spaceskip=0pt\relax}
\providecommand{\BIBentryALTinterwordstretchfactor}{4}
\providecommand{\BIBentryALTinterwordspacing}{\spaceskip=\fontdimen2\font plus
\BIBentryALTinterwordstretchfactor\fontdimen3\font minus
  \fontdimen4\font\relax}
\providecommand{\BIBforeignlanguage}[2]{{%
\expandafter\ifx\csname l@#1\endcsname\relax
\typeout{** WARNING: IEEEtran.bst: No hyphenation pattern has been}%
\typeout{** loaded for the language `#1'. Using the pattern for}%
\typeout{** the default language instead.}%
\else
\language=\csname l@#1\endcsname
\fi
#2}}
\providecommand{\BIBdecl}{\relax}
\BIBdecl

\bibitem{icp}
P.~J. Besl and N.~D. McKay, ``A method for registration of 3-d shapes,''
  \emph{IEEE Transactions on Pattern Analysis and Machine Intelligence},
  vol.~14, no.~2, pp. 239--256, Feb 1992.

\bibitem{super4pcs}
\BIBentryALTinterwordspacing
N.~Mellado, D.~Aiger, and N.~J. Mitra, ``Super 4pcs fast global pointcloud
  registration via smart indexing,'' \emph{Computer Graphics Forum}, vol.~33,
  no.~5, pp. 205--215, 2014. [Online]. Available:
  \url{http://dx.doi.org/10.1111/cgf.12446}
\BIBentrySTDinterwordspacing

\bibitem{aldoma2012our}
A.~Aldoma, F.~Tombari, R.~B. Rusu, and M.~Vincze, ``Our-cvfh--oriented, unique
  and repeatable clustered viewpoint feature histogram for object recognition
  and 6dof pose estimation,'' in \emph{Joint DAGM (German Association for
  Pattern Recognition) and OAGM Symposium}.\hskip 1em plus 0.5em minus
  0.4em\relax Springer, 2012, pp. 113--122.

\bibitem{tombari2010unique}
F.~Tombari, S.~Salti, and L.~Di~Stefano, ``Unique signatures of histograms for
  local surface description,'' in \emph{European conference on computer
  vision}.\hskip 1em plus 0.5em minus 0.4em\relax Springer, 2010, pp. 356--369.

\bibitem{Ren2015FasterRCNN}
\BIBentryALTinterwordspacing
S.~Ren, K.~He, R.~B. Girshick, and J.~Sun, ``Faster {R-CNN:} towards real-time
  object detection with region proposal networks,'' \emph{CoRR}, vol.
  abs/1506.01497, 2015. [Online]. Available:
  \url{http://arxiv.org/abs/1506.01497}
\BIBentrySTDinterwordspacing

\bibitem{Long2015FCNN}
J.~Long, E.~Shelhamer, and T.~Darrell, ``Fully convolutional networks for
  semantic segmentation,'' in \emph{2015 IEEE Conference on Computer Vision and
  Pattern Recognition (CVPR)}, June 2015, pp. 3431--3440.

\bibitem{kendall2015posenet}
A.~Kendall, M.~Grimes, and R.~Cipolla, ``Posenet: A convolutional network for
  real-time 6-dof camera relocalization,'' in \emph{Proceedings of the IEEE
  international conference on computer vision}, 2015, pp. 2938--2946.

\bibitem{puljiz2019referencing}
\BIBentryALTinterwordspacing
D.~Puljiz, K.~S. Riesterer, B.~Hein, and T.~Kr{\"o}ger, ``Referencing between a
  head-mounted device and robotic manipulators,'' in \emph{Proceedings of the
  2nd Workshop on Virtual, Mixed and Augmented Reality Human.Robot Interaction,
  HRI 2019}, 2019. [Online]. Available: \url{http://arxiv.org/abs/1904.02480}
\BIBentrySTDinterwordspacing

\bibitem{Quintero2018ARProg}
C.~P. Quintero, S.~Li, M.~K. Pan, W.~P. Chan, H.~F. M.~V. der Loos, and
  E.~Croft, ``Robot programming through augmented trajectories in augmented
  reality,'' in \emph{2018 IEEE/RSJ International Conference on Intelligent
  Robots and Systems (IROS)}, Oct 2018, pp. 1838--1844.

\bibitem{puljiz2019handguidance}
D.~{Puljiz}, E.~{St{\"o}hr}, K.~S. {Riesterer}, B.~{Hein}, and T.~{Kr{\"o}ger},
  ``{General Hand Guidance Framework using Microsoft HoloLens},'' \emph{arXiv
  e-prints}, p. arXiv:1908.04692, Aug 2019.

\bibitem{gaschler2014intuitive}
A.~Gaschler, M.~Springer, M.~Rickert, and A.~Knoll, ``Intuitive robot tasks
  with augmented reality and virtual obstacles,'' in \emph{Robotics and
  Automation (ICRA), 2014 IEEE International Conference on}.\hskip 1em plus
  0.5em minus 0.4em\relax IEEE, 2014, pp. 6026--6031.

\bibitem{weng2019refe}
T.~{Weng}, L.~{Perlmutter}, S.~{Nikolaidis}, S.~{Srinivasa}, and M.~{Cakmak},
  ``Robot object referencing through legible situated projections,'' in
  \emph{2019 International Conference on Robotics and Automation (ICRA)}, May
  2019, pp. 8004--8010.

\bibitem{chakraborti2017}
\BIBentryALTinterwordspacing
T.~Chakraborti, S.~Sreedharan, A.~Kulkarni, and S.~Kambhampati, ``Alternative
  modes of interaction in proximal human-in-the-loop operation of robots,''
  \emph{CoRR}, vol. abs/1703.08930, 2017. [Online]. Available:
  \url{http://arxiv.org/abs/1703.08930}
\BIBentrySTDinterwordspacing

\bibitem{GANeratedHands_CVPR2018}
\BIBentryALTinterwordspacing
F.~Mueller, F.~Bernard, O.~Sotnychenko, D.~Mehta, S.~Sridhar, D.~Casas, and
  C.~Theobalt, ``Ganerated hands for real-time 3d hand tracking from monocular
  rgb,'' in \emph{Proceedings of Computer Vision and Pattern Recognition
  ({CVPR})}, June 2018. [Online]. Available:
  \url{https://handtracker.mpi-inf.mpg.de/projects/GANeratedHands/}
\BIBentrySTDinterwordspacing

\bibitem{Zhang2019InteractionFusion}
\BIBentryALTinterwordspacing
H.~Zhang, Z.-H. Bo, J.-H. Yong, and F.~Xu, ``Interactionfusion: Real-time
  reconstruction of hand poses and deformable objects in hand-object
  interactions,'' \emph{ACM Trans. Graph.}, vol.~38, no.~4, pp. 48:1--48:11,
  Jul. 2019. [Online]. Available:
  \url{http://doi.acm.org/10.1145/3306346.3322998}
\BIBentrySTDinterwordspacing

\bibitem{liu2018interactive}
H.~Liu, Y.~Zhang, W.~Si, X.~Xie, Y.~Zhu, and S.-C. Zhu, ``Interactive robot
  knowledge patching using augmented reality,'' in \emph{2018 IEEE 21st
  International Conference on Robotics and Automation (ICRA)}, 2018, pp. 1947
  --1954.

\bibitem{luebbers2019augmented}
M.~B. Luebbers, C.~Brooks, M.~J. Kim, D.~Szafir, and B.~Hayes, ``Augmented
  reality interface for constrained learning from demonstration,'' in
  \emph{Proceedings of the 2nd International Workshop on Virtual, Augmented,
  and Mixed Reality for HRI (VAM-HRI)}, 2019.

\end{thebibliography}

\end{document}